\documentclass[letterpaper]{article} % DO NOT CHANGE THIS
\usepackage{aaai24}  % DO NOT CHANGE THIS
\usepackage{times}  % DO NOT CHANGE THIS
\usepackage{helvet}  % DO NOT CHANGE THIS
\usepackage{courier}  % DO NOT CHANGE THIS
\usepackage[hyphens]{url}  % DO NOT CHANGE THIS
\usepackage{graphicx} % DO NOT CHANGE THIS
\usepackage{natbib}  % DO NOT CHANGE THIS AND DO NOT ADD ANY OPTIONS TO IT
\usepackage{caption} % DO NOT CHANGE THIS AND DO NOT ADD ANY OPTIONS TO IT
\usepackage{multirow}
\usepackage{multicol}
\usepackage{subfigure}
\usepackage{amssymb}
\usepackage{booktabs}
\usepackage{hyperref}

\usepackage{algorithm}
\usepackage{algorithmic}
\usepackage{newfloat}
\usepackage{listings}
\DeclareCaptionStyle{ruled}{labelfont=normalfont,labelsep=colon,strut=off} % DO NOT CHANGE THIS
\floatstyle{ruled}
\newfloat{listing}{tb}{lst}{}
\floatname{listing}{Listing}
\title{Knowledge-Driven CoT: Exploring Faithful Reasoning in LLMs for Knowledge-intensive Question Answering}
\author{
    Keheng Wang\equalcontrib \textsuperscript{\rm 1},
    Feiyu Duan\equalcontrib \textsuperscript{\rm 1},
    Sirui Wang \textsuperscript{\rm 3},
    Peiguang Li \textsuperscript{\rm 3},
    Yunsen Xian \textsuperscript{\rm 3},
    Chuantao Yin \textsuperscript{\rm 1},
    Wenge Rong \textsuperscript{\rm 2},
    Zhang Xiong \textsuperscript{\rm 2}
}
\affiliations{
    \textsuperscript{\rm 1}Sino-French Engineering School, Beihang University\\
    \textsuperscript{\rm 2}School of Computer Science and Engineering, Beihang University\\
    \textsuperscript{\rm 3}Meituan Inc., Beijing, China\\

    wkh9575638@buaa.edu.cn,
    duanfeiyu@buaa.edu.cn,
    wangsirui@meituan.com,
    lipeiguang@meituan.com,
    xianyunsen@meituan.com,
    chuantao.yin@buaa.edu.cn,
    w.rong@buaa.edu.cn,
    xiongz@buaa.edu.cn

}

\usepackage{bibentry}
\begin{document}

\maketitle

\begin{abstract}
Equipped with Chain-of-Thought (CoT), Large language models (LLMs) have shown impressive reasoning ability in various downstream tasks. Even so, suffering from hallucinations and the inability to access external knowledge, LLMs often come with incorrect or unfaithful intermediate reasoning steps, especially in the context of answering knowledge-intensive tasks such as KBQA. To alleviate this issue, we propose a framework called Knowledge-Driven Chain-of-Thought (KD-CoT) to verify and modify reasoning traces in CoT via interaction with external knowledge, and thus overcome the hallucinations and error propagation. Concretely, we formulate the CoT rationale process of LLMs into a structured multi-round QA format. In each round, LLMs interact with a QA system that retrieves external knowledge and produce faithful reasoning traces based on retrieved precise answers. The structured CoT reasoning of LLMs is facilitated by our developed KBQA CoT collection, which serves as in-context learning demonstrations and can also be utilized as feedback augmentation to train a robust retriever. Extensive experiments on WebQSP and ComplexWebQuestion datasets demonstrate the effectiveness of proposed KD-CoT in task-solving reasoning generation, which outperforms the vanilla CoT ICL with an absolute success rate of 8.0\% and 5.1\%. Furthermore, our proposed feedback-augmented retriever outperforms the state-of-the-art baselines for retrieving knowledge, achieving significant improvement in Hit and recall performance. Our code and data are released on \href{https://github.com/AdelWang/KD-CoT/tree/main}{https://github.com/AdelWang/KD-CoT/tree/main}.

\end{abstract}

\section{Introduction}
Large language models (LLMs) pre-trained on massive language corpora have shown impressive performance in various NLP tasks \cite{brown2020language, du2022glm, touvron2023llama1}. The ability of LLMs can be further unleashed through in-context learning conditioning on a few concatenated demonstrations without task-specific training or fine-tuning. Recent works have explored LLMs' reasoning ability to tackle complex reasoning problems through prompting \cite{wei2023chainofthought, zhou2023leasttomost} and decoding \cite{wang2023selfconsistency}. 

\begin{figure}[t]
    \centering
    \includegraphics[width=1.0\linewidth]{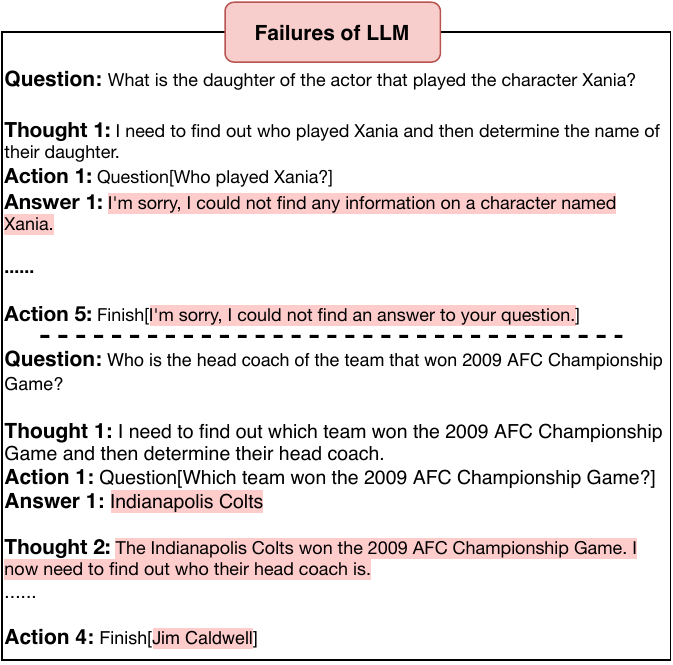}
    \caption{LLMs suffer from hallucination or inability to answer sub-questions while solving question answering tasks that require encyclopedic knowledge, resulting in erroneous subsequent reasoning and final answer. We highlight the errors with red blocks.}
    % \vspace{-3mm}
    \label{failure}
\end{figure}

Despite advancements, LLMs still encounter hallucinations or lack of knowledge while solving knowledge-intensive tasks. As shown in Figure \ref{failure}, both these failures will lead to erroneous subsequent reasoning steps and incorrect final answers. Previous work prompts LLMs to generate structured Chain-of-Thought (CoT) by searching relevant information from the Web \cite{yao2023react}, or verifies the intermediate reasoning through an additional verification system and returns the input to LLMs to re-generate the rationale \cite{wang2023boosting}. 
% However, none of them address the hallucination problem in complex multi-hop question scenarios that requires encyclopedic knowledge. 
However, the problem of hallucinations in complex multi-hop problem scenarios is still understudied.

A naive method is to directly input numerous contextual knowledge into LLMs, converting the question answering task into reading comprehension. However, ensuring comprehensive knowledge coverage necessitates a large amount of context, making it difficult for LLMs to fully understand \cite{liu2023lost}. Considering that current state-of-the-art methods leverage a retrieve-then-read pipeline that retrieves external knowledge and returns a knowledge-related answer for solving knowledge-intensive Question Answering tasks, we can address the aforementioned issue by applying such a paradigm. Recent work has either enhanced the retriever's capability to retrieve external knowledge \cite{izacard2021contriever, chuang2023expand} or improved the reader's ability to extract answers from the retrieved knowledge \cite{yu2023decaf, yu2022kgfid}. Other works focus on multi-hop Knowledge Base Question Answering (KBQA) by leveraging intermediate supervision signals \cite{He_2021} or decomposing the question into several sub-questions \cite{sun2020sparqa, khot2021text}. Nevertheless, few works leverage the understanding and reasoning capabilities of LLMs to address complex multi-hop KBQA tasks, as well as investigate the problem of model hallucination for encyclopedic knowledge. Moreover, there is a dearth study on how to boost faithful reasoning by leveraging external knowledge to improve the intermediate reasoning traces.

To address the above issue, we propose a Knowledge-Driven Chain-of-Thought (KD-CoT), an interactive framework that utilizes a QA system to access external knowledge and provide high-quality answers to LLMs for solving knowledge-intensive KBQA tasks.
% 这里要详细讲吗，感觉这些应该放在method里面，ReAct也是笼统地讲一下，再展开讲篇幅好像多了。Specifically, we formulate a structured multi-round QA format,  train a retriever-then-reader pipeline that effectively retrieves external knowledge and returns a knowledge-driven answer. Then the 
KD-CoT is designed to encourage LLMs to generate verbal reasoning traces, facilitating dynamic reasoning that can verify and adjust intermediate reasoning steps by accessing external knowledge. We also create a KBQA CoT collection that can be applied to perform ICL, and train a robust retriever utilizing the constructed collection.

Our main contributions can be summarized as follows:
\begin{itemize}
    \item We present a KBQA CoT collection by prompting LLMs, which could be used for fine-tuning smaller LMs to acquire CoT reasoning ability and be applied to perform ICL.
    \item We propose a retriever-reader-verifier QA system to access external knowledge and interact with LLM. We leverage the constructed CoT collection as feedback augmentation to train a robust retriever, achieving significant improvement in Hit scores on WebQSP and CWQ.
    \item We introduce KD-CoT, a Knowledge-Driven Chain-of-Thought framework to improve the reasoning performance of large language models. Experimental results demonstrate the effectiveness of our proposed framework, achieving 8.0 and 5.1 Hit@1 improvement on WebQSP and CWQ compared to the vanilla ICL method.

% （3）我们提出了一个Verifier-then-reasoning框架，即在推理的过程中进行交互，通过retrieve-then-read的方式为每一个中间子问题生成答案候选，并用一个verifier进行筛选，以此提高推理过程的可靠性。实验结果表明我们提出的框架能够一定程度上修正因幻觉产生的中间推理不可靠问题，进一步提升CoT QA的表现。
\end{itemize}

\section{Methodology}
In this section, we first present the procedure for constructing the CoT Collection. Then we introduce the Knowledge-Driven CoT framework, which encompasses the implementation of the interaction and the training of the external QA system.
%This framework aims to enhance the accuracy of reasoning sub-steps and bolster the dependability of the entire reasoning process.

\begin{figure}[t]
    \centering
    \includegraphics[width=1.0\linewidth]{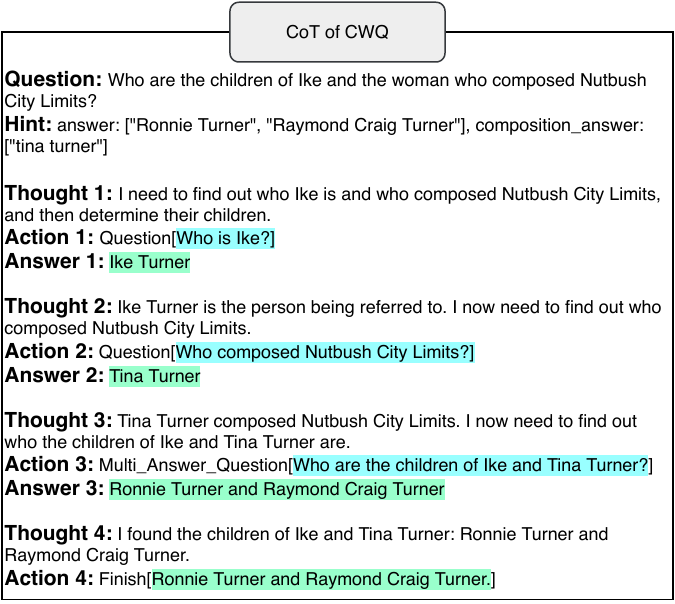}
    \caption{Illustration of manually written demonstrations. Blue and Green parts denote the sub-question and sub-answer generated by LLM, respectively.}
    \label{illustration-demon}
    % \vspace{-3mm}
\end{figure}

\begin{figure*}[t]
    \centering
    \includegraphics[width=1.0\linewidth]{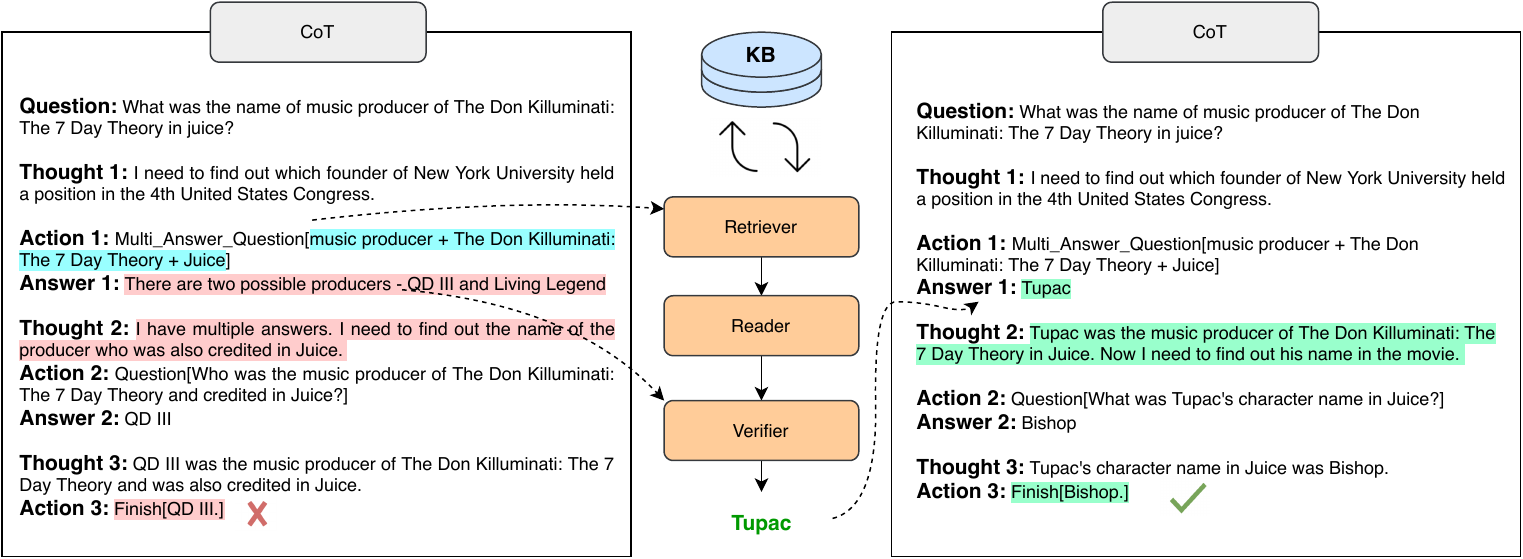}
    \caption{The overall framework of Knowledge-Driven CoT, including a prompted large model and a QA system that accesses external knowledge. By modifying sub-answers of intermediate questions, LLM can generate more faithful subsequent inference steps, which lead to correct final answers. Blue, Green, and Red blocks represent the sub-question fed to QA system, correct/incorrect reasoning and answers, respectively}
    \label{framework}
    \vspace{-2mm}
\end{figure*}

\subsection{CoT Collection}
\label{subsection-cot}
% Despite the effectiveness of rationale data for CoT fine-tuning, it is very difficult to construct high-quality knowledge-intensive rationales due to the difficulty in gathering human-authored rationales \cite{kim2023cotever} and the hallucinations of large models \cite{bang2023multitask}. In this section, we provide a detailed description of the procedure used to construct our Knowledge-intensive rationales.
The rationale data for CoT fine-tuning has shown great value, while constructing such high-quality rationales is quite challenging due to the difficulty in gathering human-authored rationales \cite{kim2023cotever} and the hallucinations of large models \cite{bang2023multitask}. Here we provide a detailed description of how we construct knowledge-intensive rationales with the help of LLMs.

It has been proven that LLMs perform ICL conditioning on demonstrations have a better understanding of the task and generate more accurate responses \cite{brown2020language}. Inspired by this, we assign the demonstrations every time request LLMs. To obtain the demand demonstrations, we first manually write several accurate CoT demonstrations as the anchor set, and then we employ an iterative algorithm to construct our full collection. In each iteration, we choose the candidate in the current collection that holds the highest cosine similarity with the question in the training set to serve as the demonstration. We utilize RoBERTa \cite{reimers-2019-sentence-bert}\footnote{Downloaded from \href{https://huggingface.co/sentence-transformers/roberta-large-nli-stsb-mean-tokens}{Sentence Transformers; Roberta-large-nli-stsb}} to embed questions. Next, we request ChatGPT\footnote{\href{https://chat.openai.com/}{OpenAI gpt-3.5-turbo}} to generate the structured CoT, and append generated results "Finish" with the correct answer to the collection. The construction details are referred to Algorithm \ref{alg-cot}. Notably, we observe that concatenating the ground truth answer and the composition answer (if have) as "Hint" before the rationale can greatly improve the efficiency of collection construction, so the final demonstration is presented in the format of \textless Question, Hint, CoT\textgreater as illustrated in Figure \ref{illustration-demon}. 

% Next, we prompt ChatGPT\footnote{{OpenAI gpt-3.5-turbo}} to generate the structured CoT, and append generated results "Finish" with the correct answer to the collection. Please refer to Algorithm \ref{alg-cot} for more details.

\begin{algorithm}[t]
	\caption{Construct CoT Collection} 
	\label{alg-cot} 
	\begin{algorithmic}
		\REQUIRE Human-annotated demonstrations, $D_{h}$
            \REQUIRE A fixed human-annotated instruction, $I$
            \REQUIRE Question-Answer training set, $Q, \ A$
            \REQUIRE Large language model, $LLM$
            \REQUIRE Demonstration selection pool, $P$
            \STATE $P \gets D_{h}$
            \STATE $iteration \gets 0$
		\WHILE {$Q$ is not empty and $iteration < 5$} 
		\STATE $Demons \gets SimilaritySelection(Q, \ P)$
            \STATE $Inputs \gets Concat(I, \ Demons, \ Q)$
            \STATE $Outputs \gets LLM(inputs)$ 
            \STATE $Constructed \gets AnswerMatch(outputs, \ A)$ 
            \STATE $P \gets Extend(P, \ Constructed)$ 
            \STATE $Q \gets Q \ \backslash \ A$ 
            \STATE $iteration \gets iteration + 1$ 
		\ENDWHILE
            \STATE \textbf{return} $P$
	\end{algorithmic} 
\end{algorithm}
\subsection{Knowledge-Driven CoT}
Due to hallucinations and the inability to access external knowledge, LLMs struggle to generate faithful reasoning steps for knowledge-intensive QA tasks. To address this issue, we propose Knowledge-Driven Chain-of-Thought Reasoning (KD-CoT), which incorporates a QA system to interact with LLMs. The overall framework of our proposed KD-CoT is shown in Figure \ref{framework}.

% For each question in the test set, we select the instance with the highest cosine similarity from the collection, and utilize its rationale as the demonstration to perform one-shot ICL\footnote{Increasing the number of ICL demonstrations will improve the performance of LLMs, but also much more costly. We only concatenate a single demonstration for CoT ICL if not specified.}. We then extract the intermediate sub-question as the input of the QA system, which is comprised of a retrieve-then-read pipeline and an answer verifier. The former module retrieves external knowledge and proposes a candidate answer based on the retrieved information, while the latter chooses between the original sub-answers generated by LLM and the proposed candidate answer. We repeat the above interaction until the CoT is finished. Note that our motivation is to supervise the intermediate reasoning of LLM and not to alter the ultimate answer. Therefore, we restrict our interaction with the external QA system to sub-questions leading up to the final \textbf{Action}. For example, \textbf{Action 4} "Finish" the entire CoT in Figure \ref{illustration-demon}, we solely verify the sub-answers of \textbf{Action 1}, \textbf{2} and \textbf{3} iteratively.
For each question in the test set, we select the instance with the highest cosine similarity from the collection, and utilize its rationale as the demonstration to perform one-shot ICL\footnote{Increasing the number of ICL demonstrations will improve the performance of LLMs, but also much more costly. We only concatenate a single demonstration for CoT-ICL if not specified.}. Then the extracted intermediate sub-question is taken as the input of the QA system to perform interaction, which is comprised of a retrieve-then-read pipeline and an answer verifier. The former module retrieves external knowledge and proposes a candidate answer based on the retrieved information, while the latter chooses between the original sub-answers generated by LLM and the proposed candidate answer. We repeat the above interaction until the CoT is finished. Note that our motivation is to supervise the intermediate reasoning of LLM and not to alter the ultimate answer. Therefore, we restrict our interaction with the external QA system to sub-questions leading up to the final \textbf{Action}. For example, \textbf{Action 4} "Finish" the entire CoT in Figure \ref{illustration-demon}, we solely verify the sub-answers of \textbf{Action 1}, \textbf{2} and \textbf{3} iteratively.

%% todo 0816 要不要修改这里，因为sota分数更好，用sota模型得到的分数还不一定比直接过 retrieve-then-read pipeline 分数高，感觉需要换套说辞。
To ensure the accuracy of the intermediate reasoning steps, it is crucial to have a strong QA system that can effectively access external knowledge and generate highly precise answers. We then introduce how to train our retrieve-then-read pipeline and the verifier.

\noindent \textbf{KB Linearization}
We aim to interact with both structural (KBs) and unstructured (Wikipedia) external knowledge. However, directly retrieving information from KBs is non-trivial due to its large scale and complication with semantics and structure. To address this issue, we simply follow the linearization method proposed in \cite{yu2023decaf} to process Freebase KB data \cite{10.1145/1376616.1376746} into unstructured text. Given a head entity, we extract its 1-hop subgraph and concatenate the KB triplets with spaces, then group the entire subgraph into a single passage. For example (music recording, releases, Palavras de Guerra Ao Vivo) and (music recording, artist, Olívia Hime) will be processed into "music recording releases Palavras de Guerra Ao Vivo. music recording artist Olívia Hime".

After pre-processing, we concatenate Wikipedia passages with KB passages to perform knowledge retrieval.

\noindent \textbf{Feedback-Augmented Retriever}
%% todo
To align with previous work \cite{oguz2022unikqa, yu2023decaf}, we apply Dense Passage Retrieval (DPR) \cite{karpukhin2020dense} as the model architecture of the retrieval system. To obtain a robust retriever, we propose to utilize the constructed CoT as feedback to identify relevant passages. Specifically, we extract the last reasoning sub-question from the CoT rationale as the query augmentation and concatenate it with the original question. To identify positive and negative passages relevant to the question, we further concatenate the augmented query with the answers\footnote{Query mentioned below stands for \textless question, rationale question\textgreater, and BM25 searches the relevant passages on \textless question, rationale question, answers\textgreater}, and use the BM25 algorithm to extract the top 100 related passages. We identify passages that contain entities present in both the query and answer as positive examples, while passages that only contain the answer or query entities are considered hard negatives. If no co-occurrence passage is found, we follow the primary settings in the original paper and use the passage containing only the answer as positive to ensure the recall rate of the multi-answer question. We utilize Spacy\footnote{\href{https://spacy.io/}{https://spacy.io/}} to recognize named entities in the query. Note that the feedback of LLM is only used for identifying positive/negative passages, we use the original questions to train our DPR model.

\begin{figure}[t]
    \centering
    \includegraphics[width=1.0\linewidth]{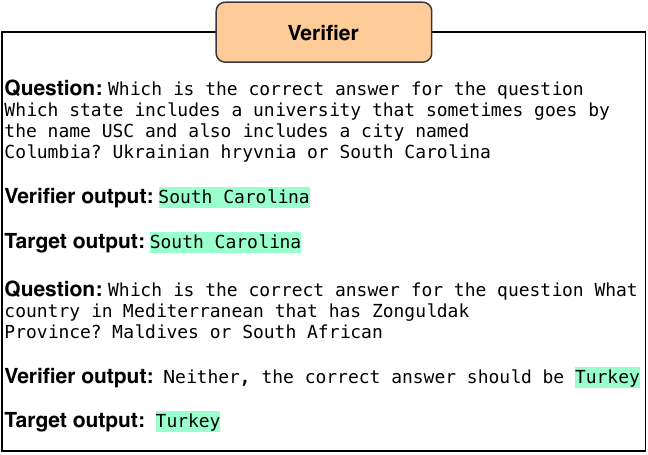}
    \caption{Illustrations of input-output pairs of the verifier. We highlight the correct answers with green blocks.}
    \label{verifier-case}
    \vspace{-3mm}
\end{figure}

\noindent \textbf{Fuse-in-Decoder Reader}
For our reader, we use the mainstream Fuse-in-Decoder architecture \cite{izacard2021leveraging} to train a Transformer \cite{vaswani2023attention} model. Specifically, given a question q and its top-N relevant passages P, the FiD reader first separately encodes each passage $p_{q_{i}}$ concatenated with $q$:
\begin{equation}
    \centering
    P_{i} = \mbox{Encoder}(\mbox{Concat}[q, \ p_{q_{i}}]) \in \mathbb{R}^{L\times H}
\end{equation}

Where L, H represent sequence length and hidden size, respectively. Then the token embeddings of all passages output from the encoder are concatenated and fed to the decoder to generate the final answer. Different from previous work, we employ all answers as training targets instead of selecting one randomly.
\begin{equation}
    \centering
    A = \mbox{Decoder}(\mbox{Concat}[P_{1}, P_{2}, ..., P_{N}])
\end{equation}

\noindent \textbf{Verifier}
We train a Llama2-7b \cite{touvron2023llama} with Parameter-Efficient-Fine-Tuning (PEFT) on the original KBQA training set as our verifier. During inference, the model takes the original sub-answers generated by LLM and the candidate answers generated by the retrieve-then-read pipeline as input, and outputs its preferred one. If neither answer is selected, the verifier will generate a new answer. We illustrate several examples in Figure \ref{verifier-case}.

If not specified, greedy decoding is used for Reader and Verifier during inference.

\section{Experiment}
\subsection{Experiment Settings}
For our main experiment, we use ChatGPT as our backbone "thinker" to interact with an external QA system, which is denoted as "LLM" in the subsequent sections of this paper. The QA system includes a BERT-base \cite{devlin2019bert} retriever, a T5-large \cite{raffel2020exploring} reader, and a Llama2-7b verifier fine-tuned with LoRA \cite{hu2021lora} \footnote{We use the checkpoints downloaded from \href{https://huggingface.co/models}{Huggingface.}}. We prompt LLM to perform structured multi-round QA reasoning with demonstrations selected from our constructed CoT collection.

We train our retriever on a merged dataset of WebQSP and CWQ for saving the cost of embedding massive knowledge and use the same DPR architecture in the original paper \cite{karpukhin2020dense}. The number of retrieved passages is 100 if not specified. The reader is also trained on the merged dataset to effectively tackle both single-hop and multi-hop question scenarios.

To show the effectiveness and correctness of our CoT collection, we also conduct experiments that involve fine-tuning smaller models on the constructed rationale data. We use Flan-T5-3B \cite{chung2022scaling} and T5-3B \cite{raffel2020exploring} as the base models and compare the results with text-to-text QA fine-tuning. To indicate the CoT paradigm that generates both rationale and answers, we incorporate a trigger phrase "Let's think step by step" into the sequence during training and evaluation. 
%To investigate the effectiveness of our proposed DPR, we evaluate our model based on metrics Hits@N and Recall@N, where Hits@N focus

\noindent \textbf{Dataset} \quad We evaluate KD-CoT on two KBQA datasets: WebQSP \cite{yih-etal-2016-value} and ComplexWebQuestions (CWQ) \cite{talmor-berant-2018-web}. 
%Specifically, WebQSP stands for few-hop QA, while CWQ refers to a more complex multi-hop scenario. 
We use the original datasets to train our external QA system, and use the constructed CoT collection to apply ICL on ChatGPT. The data statistics are shown in Table \ref{data statistic}.

\begin{table}[ht]
    \centering
    \scalebox{0.9}{
        \begin{tabular}{c|cc|cc}
            \toprule
            \multirow{2}{*}{\# Data} & \multicolumn{2}{c|}{\textbf{WebQSP}} & \multicolumn{2}{c}{\textbf{CWQ}} \\
            \cline{2-5}
            \multirow{2}{*}{} & train & test & train & test \\
            \hline
            original & 3098 & 1639 & 27625 & 3519 \\
            \hline
            CoT collection & 2888 & 1639 & 26695 & 3519 \\
            \bottomrule
        \end{tabular}
    }
    \caption{Data statistics of original datasets and our CoT collections. After collection construction, we obtained 2888 and 26695 rationale data for WebQSP and CWQ, respectively.}
    \label{data statistic}
    % \vspace{-3mm}
\end{table}

\noindent \textbf{Evaluation metric} \quad Following previous work, we evaluate our model based on metrics Hits@1 and F1, where Hits@1 focuses on the single top-ranked answer while F1 considers coverage of all the answers. To account for the fact that it's difficult to extract the desired answers from LLM's output, we adjust our evaluation criteria. We deem the generated results to be correct if they contain the ground truth answer.

%% todo, 跑出结果后看分数写analyse

\subsection{Konwledge-Driven CoT Results}
% 先讲我们的方法的好，即相对于原始的react分数 / multi-QA pairs 有很大提升
Table \ref{main result} reports the performance of our proposed KD-CoT. The results show that KD-CoT outperforms vanilla CoT ICL by a significant margin of 8.0 and 5.1 points on WebQSP and CWQ, respectively. This highlights the effectiveness of interacting with the external QA system, as it enables the LLM to generate more accurate intermediate reasoning steps, leading to more precise final answers.
% We also investigate the performance of the LLM when incorporating knowledge retrieved by DPR and other Q-A pairs as context to apply ICL, which are referred to as LLM$_{Retrieval}$ and LLM$_{QA \ pairs}$, respectively. To align with one-shot CoT ICL, we restrict the input length and concatenate only 4 passages/QA pairs as the context that fed into LLM and the results are shown in Table \ref{main result}. It can be observed that the LLM's performance is inferior to vanilla CoT ICL under these two settings, especially in complex multi-hop problem scenarios. This indicates that concatenating a small amount of knowledge or QA pairs as a demonstration does not bring gains to the model's performance.

\begin{table}[t]
    \scalebox{0.95}{
        \begin{tabular}{l|c|c|c}
        \toprule 
        \multirow{2}{*}{\textbf{Method $\backslash$ Dataset}} & \multicolumn{2}{c|}{\textbf{WebQSP}} & \textbf{CWQ} \\
        \cline{2-4}
        \multirow{2}{*}{} & Hit@1 & F1 & Hit@1 \\
        \hline
        UnikQA \cite{oguz2022unikqa} &   79.1    & - & - \\
        DeCAF \cite{yu2023decaf} & 80.7 & 77.1 & 67.0 \\
        DeCAF$_{w / o \ LF}$ \cite{yu2023decaf} & 74.2 & 49.5 & 47.9 \\
        \hline
        Our$_{\ Retrieve-then-read}$ & 73.7 & 50.2 & 50.5 \\
        \hline
        LLM$_{\ Retrieval \ 4-passages}$ & 52.4 & 38.2 & 26.9 \\
        LLM$_{\ QA \ pairs \ 4-shot}$ & 53.2 & 39.2 & 42.2 \\
        LLM$_{\ CoT \ fixed}$ & 50.3 & 37.8 & 34.0 \\
        LLM$_{\ QA-CoT \ fixed}$ & 56.6 & 42.5 & 42.4 \\
        LLM$_{\ QA-CoT \ selected}$ & 60.6 & 47.8 & 50.6 \\
        \hline
        KD-CoT & 68.6 & 52.5 & 55.7 \\
        KD-CoT$_{w/o \ Retrieve-then-read}$ & 66.8 & 49.4 & 49.2 \\
        KD-CoT$_{w/o \ Verifier}$ & 59.9 & 47.6 & 49.2 \\
        \bottomrule
        \end{tabular}
    }
    \caption{Experimental results on WebQSP and CWQ. KD-CoT significantly outperforms the vanilla CoT ICL.}
    \label{main result}
\end{table}

\begin{figure}[ht]
    % \vspace{-2mm}
    \centering
    \begin{minipage}[b]{1.0\linewidth}
        \includegraphics[width=0.48\linewidth]{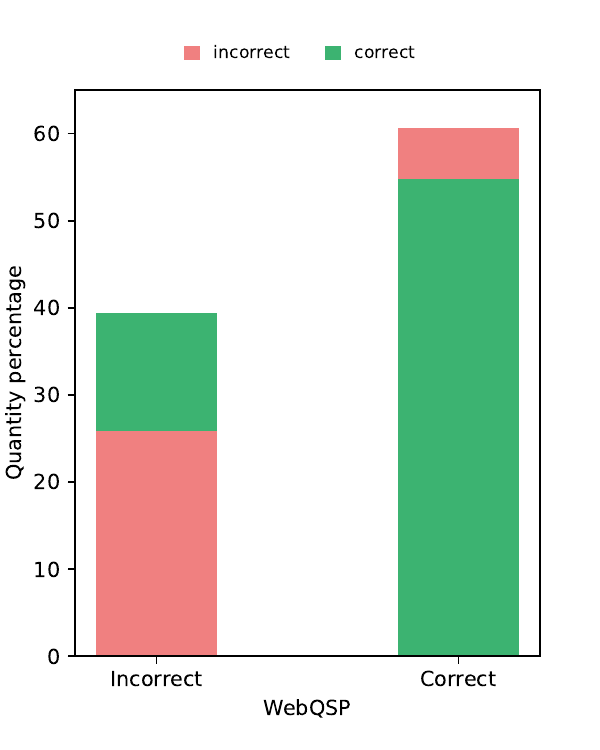}
        % \hspace{-2mm}
        \includegraphics[width=0.48\linewidth]{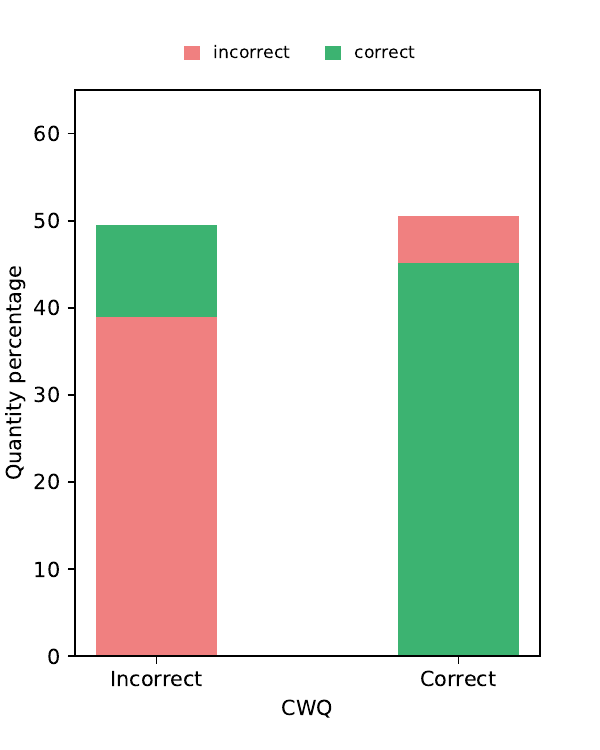}
    \end{minipage}
    \caption{Quantity percentage of questions answered correctly/incorrectly by the LLM. The horizontal axis represents the state before passing through the interaction framework. Red and Green blocks represent the proportion of questions answered correctly/incorrectly after the entire interaction.}
    \label{correctness}
    \vspace{-3mm}
\end{figure}

To further demonstrate that our process of verifying and correcting sub-answers can lead to faithful reasoning so as to rectify previously incorrect responses, we tally the alterations in the count of correct and incorrect answers before and after undergoing our interactive framework. The results are shown in Figure \ref{correctness}. On WebQSP and CWQ, we correct 13.5\% and 10.6\% of questions that were incorrectly answered previously, while only 5.8\% and 5.4\% are modified to be incorrect.

% However, the inference performance of LLM still underperforms the current state-of-the-art results after answer correction, indicating that 

Despite its efficiency in extracting sub-questions to interact with the external QA system, ReAct format CoT also reduces its flexibility in formulating reasoning steps due to the structural constraint \cite{yao2023react}. Once the output generated is inadequately structured and unable to extract sub-questions, we consider it a failure in answering. Consequently, the capability of LLM might be underestimated.
\begin{figure*}[t]
    % \vspace{-2mm}
    \centering
    \subfigure[LLM performance after each iteration of interaction. The highest performance is achieved in the first iteration for WebQSP and the last iteration for CWQ, as WebQSP is primarily comprised of single-hop questions, whereas CWQ contains more complex multi-hop questions.]{ \label{iteration score}
    \centering
    \begin{minipage}[b]{1.0\linewidth}
        \includegraphics[width=0.48\linewidth]{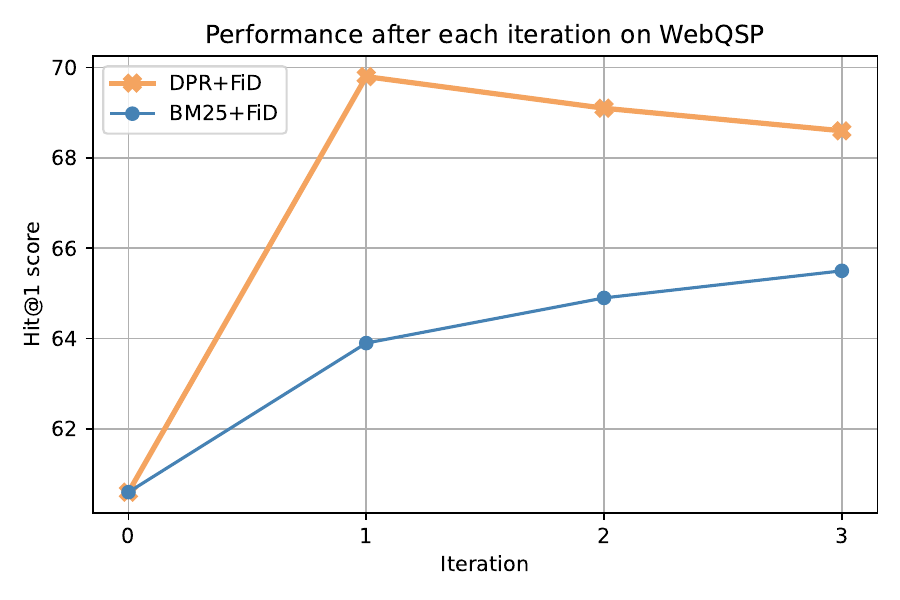}
        \includegraphics[width=0.48\linewidth]{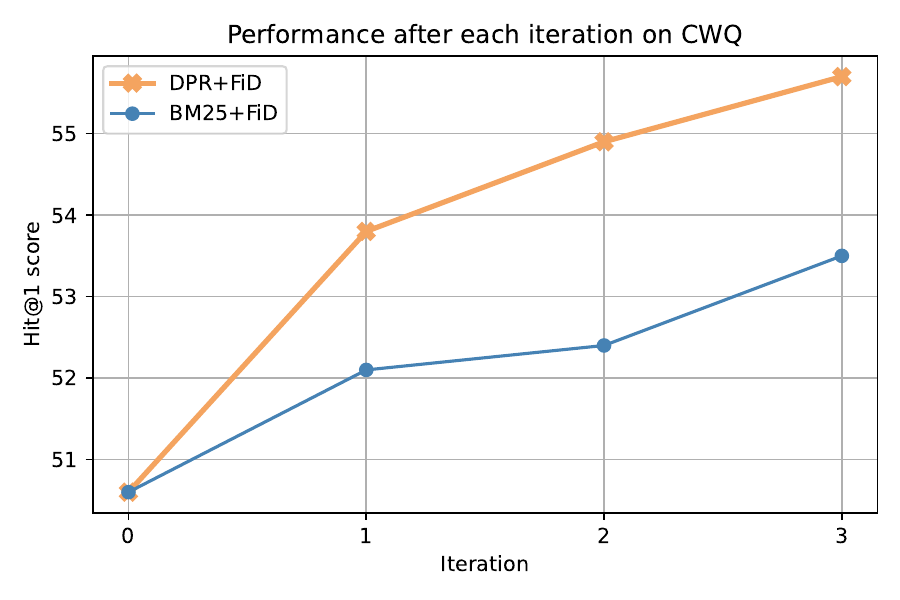}
    \end{minipage}} \\
    % \vspace{-2mm}
    \subfigure[Answer source during each iteration. In most cases, the verifier prefers sub-answers output by ChatGPT, about half of the sub-answers are modified by the external QA system in each iteration.]{ \label{iteration source}
    \centering
    \begin{minipage}[b]{1.0\linewidth}
        \includegraphics[width=0.5\linewidth]{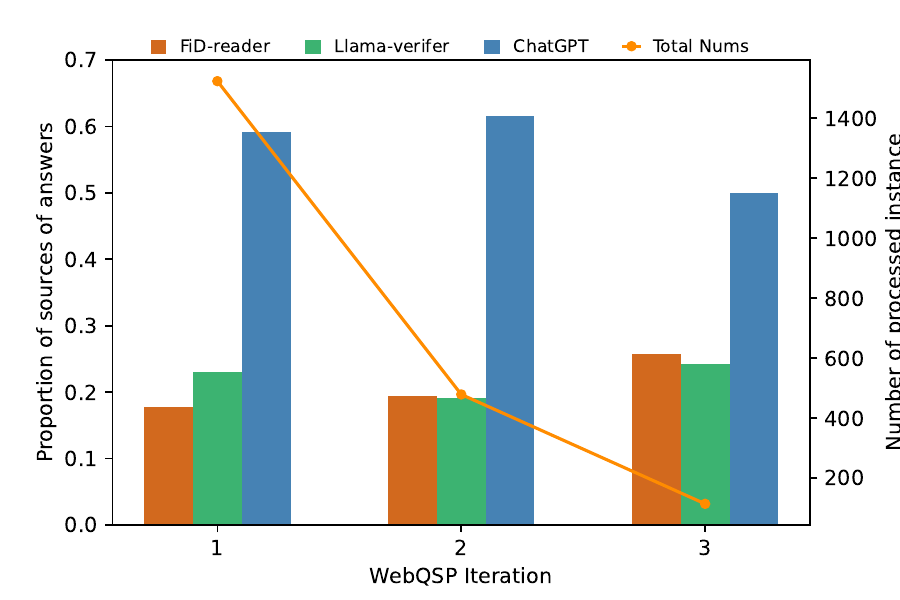}
        \hspace{-3mm}
        \includegraphics[width=0.5\linewidth]{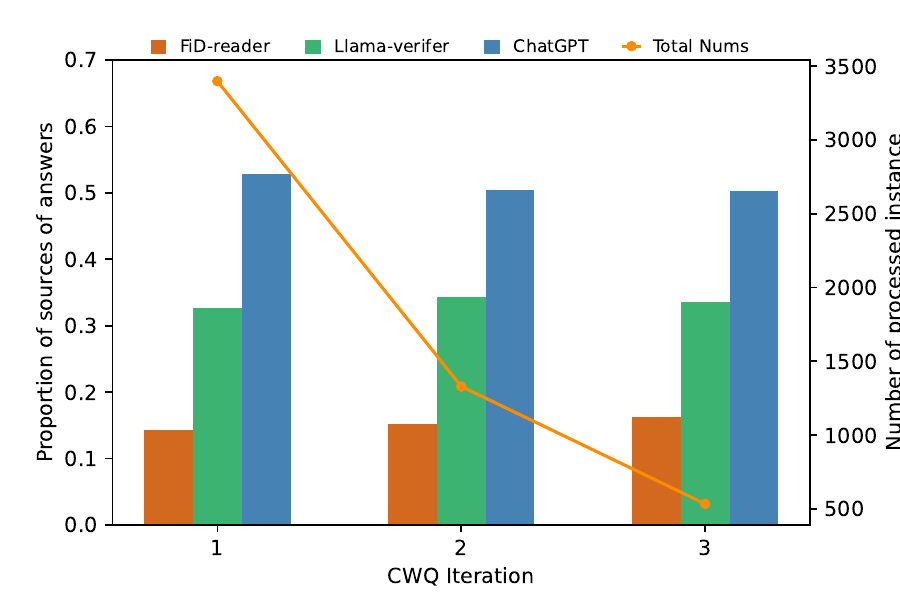}
    \end{minipage}}
    
    \caption{a) LLM performance after each iteration of interaction. b) Answer source during each iteration.}
    \label{benefits}
    \vspace{-2mm}
\end{figure*}

\subsection{Retrieval Results}
We evaluate the effectiveness of our feedback-augmented DPR in Table \ref{retrieval results}. It can be seen that our proposed FBA-DPR significantly outperforms previous results in \citet{yu2023decaf} on both WebQSP and CWQ, achieving 3.8 and 7.8 points of improvement on Hit@100.

\begin{table}[t]
    \centering
    \scalebox{0.85}{
        \begin{tabular}{c|cc|cc}
            \toprule
            \multirow{2}{*}{\textbf{Method}} & \multicolumn{2}{c|}{\textbf{WebQSP}} & \multicolumn{2}{c}{\textbf{CWQ}} \\
            \cline{2-5}
            \multirow{2}{*}{} & H / R@20 & H / R@100 & H / R @20 & H / R@100 \\
            \hline
            BM25 & 66.8 / 49.8 & 83.8 / 69.8 & 47.8 / 42.7 & 65.4 / 59.3 \\
            DPR & - / - & 91.6 / 80.6 & - / - & 71.4 / 65.6 \\
            \hline
            FBA-DPR & 89.0 / 75.6 & 95.4 / 88.4 & 68.7 / 62.5 & 81.3 / 76.5 \\
            $_{\ {} \ \ }$ $_{w/o \ wiki}$ & 88.9 / 74.2 & 94.8 / 86.3 & 65.0 / 58.5 & 77.8 / 72.6 \\
            \bottomrule
        \end{tabular}
    }
    \caption{Retrieval results on WebQSP and CWQ. H@N and R@N stand for the answer hits rate and recall rate of Top-N retrieved passages, respectively. DPR results are copied from \citet{yu2023decaf}, BM25 and FBA-DPR results are obtained in our setting.}
    \label{retrieval results}
    % \vspace{-3mm}
\end{table}

\subsection{CoT Fine-tuning Results}
We conduct experiments under two different settings: \textbf{Direct Fine-tuning} and \textbf{CoT Fine-tuning}. For \textbf{Direct Fine-tuning} we train the model on the original QA pairs, which takes the questions as input and directly generates the answers. For \textbf{CoT Fine-tuning} we use the trigger "Let's think step by step" concatenated with the questions, and train the model to output rationales and final answers. The results are reported in Table \ref{cot-finetuning}.

\begin{table}[ht]
    \centering
    \scalebox{0.9}{
        \begin{tabular}{c|cc|cc}
            \toprule
            \multirow{2}{*}{\textbf{Model} $\backslash$ \textbf{Dataset}} & \multicolumn{2}{c|}{\textbf{WebQSP}} & \multicolumn{2}{c}{\textbf{CWQ}} \\
            \cline{2-5}
            \multirow{2}{*}{} & Direct & CoT & Direct & CoT \\
            \hline
            T5-3B & 40.8 & 41.9 & 39.4 & 38.6 \\
            FlanT5-3B & 46.1 & 47.0 & 50.5 & 43.8 \\
            \hline
            Llama2-7B-LoRA & 63.8 & 64.1 & 48.4 & 45.1 \\
            \bottomrule
        \end{tabular}
    }
    \caption{Comparison of Direct Fine-tuning and CoT Fine-tuning. Hits@1 score is reported.}
    \label{cot-finetuning}
\end{table}

We observe that fine-tuning LMs with CoT rationales slightly outperforms Direct Fine-tuning for solving simpler questions. In complex multi-hop question scenarios, CoT fine-tuning brings negative gains. This might be because 1) The reasoning procedure of the original LM differs from that of the CoT collection. Fine-tuning the LM may potentially disrupt the original knowledge, resulting in a degradation in performance; 2) LMs still struggle to generate faithful multi-step reasoning even fine-tuned with CoT when solving knowledge-intensive tasks. However, the models still achieve competitive results with Direct Fine-tuning, demonstrating the correctness of our constructed collection.

\subsection{Analysis \& Ablation Study}
This section aims to address the following question through analysis and ablation experiments.

\noindent \textbf{Benefits of structured CoT and CoT collection?} \\
To discuss the benefits of rationale in the form of structured multi-round QA, and to highlight the significance of the CoT collection we construct in performing ICL, we evaluate the following model settings, with the name corresponding to the rows in Table \ref{main result}:
\begin{itemize}
    \item LLM$_{\ Retrieval \ 4-passages}$ We roughly concatenate the top-ranked retrieved passages with the question as input and instruct LLM to answer the question.

    \item LLM$_{\ QA \ pairs \ 4-shot}$ We utilize other QA pairs with the highest cosine similarity to the target question as the demonstrations to perform ICL.

    \item LLM$_{\ CoT \ fixed}$ We manually design unstructured rationales aligned with the content of our structured CoT, and utilize them as demonstrations to prompt LLM. The in-context demonstration is selected within human-annotated unstructured rationales.

    \item LLM$_{\ QA-CoT \ fixed}$ The in-context demonstration is selected within human-annotated structured rationales.
    
    \item LLM$_{\ QA-CoT \ selected}$ The in-context demonstration is selected within our constructed CoT collection.
\end{itemize}
To align with one-shot CoT ICL, we restrict the input length and concatenate only 4 passages/QA pairs as the context that fed into LLM. Experimental results are shown in Table \ref{main result}. We observe that LLM achieves superior performance when utilizing structured rationale as the demonstration, outperforming other ICL methods. This suggests that our proposed multi-round QA format rationale is more effective in unleashing LLM's reasoning capability. Directly concatenating the retrieved knowledge does not have a positive contribution to the model's ability, especially in complex multi-hop question scenarios, and it performs the worst among all ICL methods. By employing our constructed CoT collection, we further improve the LLM's ability, highlighting the effectiveness and necessity of the collection construction.

\noindent \textbf{Benefits of QA system?} \\
To assess the effectiveness of the QA system, we conduct two supplementary experiments where we remove the retrieve-then-read pipeline and the verifier separately. The results are reported in Table \ref{main result}. We observe that the performance degrades when the retrieve-then-read pipeline is removed, showing the importance of accessing external knowledge for precise sub-answers generation. Moreover, the performance without the verifier underperforms our KD-CoT setting, showing that in certain cases the sub-answers generated by the LLM are superior. Further combining the output of the reader and LLM to generate better answers is important for improving performance.

We further investigate the performance gain after each iteration and count the source of the modified answer. The results are shown in Figure \ref{benefits}. As can be seen in Figure \ref{iteration score}, the highest performance of LLM is achieved in the first iteration for WebQSP and the last iteration for CWQ, as WebQSP is primarily comprised of single-hop questions, whereas CWQ contains more complex multi-hop questions. We also observe that LLM tends to produce redundant inference steps despite being able to answer questions within two hops of reasoning. This leads to the necessity of second and third iterations to terminate the CoT while solving WebQSP questions. An extra Halter \cite{creswell2022faithful} to determine whether the current reasoning step can answer the questions can be a possible method for solving this issue. As the number of iterations increases, the performance on the CWQ dataset also improves. This suggests that our interaction framework can assist the model in better reasoning for complex multi-hop questions.

Figure \ref{iteration source} shows the source of modified answers. In most cases, the verifier will keep the original sub-answers generated by ChatGPT, about half of the sub-answers are modified and fed to the next iteration. This implies that a robust reader capable of producing varied and accurate answers is crucial in fully unleashing the potential of LLM.

% \noindent \textbf{Performance on smaller "thinker"?} \\
% CoT fine-tuning on our constructed collection enables models to perform knowledge-intensive reasoning. However, models still struggle with hallucination and consequently exhibit subpar performance in complex multi-hop question scenarios. We investigate whether external QA systems can enhance the CoT inference of these relatively small models by providing more precise answers.

% \noindent \textbf{Does our DPR effectively retrieves useful knowledge?}

\section{Related Work}

\subsection{Chain-of-thought Prompting}
With the massive increase of model parameters and training data, models begin to emerge powerful reasoning capabilities \cite{wei2022emergent}. Inspired by this performance breakthrough, \cite{wei2023chainofthought} proposes a gradient-free method of chain-of-thought (CoT) prompting, which allows models to give reliable answers after thinking and interpreting. Based on this research, several studies have been conducted to improve the effectiveness. For example, some studies focus on how to make LLM generate more accurate and reliable chains of thought \cite{he2022rethinking, wang2023selfconsistency, lyu2023faithful}, while some works investigate more efficient ways of generating chains of thought to unleash the potential of LLM reasoning \cite{creswell2022selectioninference, zhou2023leasttomost, jin2023tabcot}. As LLMs are confined to the knowledge learned from the training corpus, extensive efforts have been made recently to facilitate LLMs in dynamically interacting with the real world to obtain the information required for model reasoning. \cite{yao2023react, press2023measuring, liu2023webglm, peng2023check}. For instance, ReAct \cite{yao2023react} suggests utilizing a blend of reasoning and action to allow LLMs to acquire background knowledge from Wikipedia and solve problems automatically, thus mitigating the hallucination issue of LLMs. However, they simply concatenate the statements found on the wiki page to the end of the sub-question without performing more accurate retrieval. In contrast, we make use of a more advanced retriever-reader pipeline to furnish the model with more accurate and targeted knowledge.

\subsection{Knowledge Base Question Answering}
Knowledge base question answering (KBQA) has been a popular research topic in recent years, and a series of approaches have been suggested to enhance the efficiency of Knowledge base QA systems. The retriever-reader pipeline is a commonly employed technique, where the retriever extracts the most pertinent corpus from a knowledge base based on the question, and the reader produces the ultimate answer by utilizing the question and the corpus retrieved. Hence, certain studies concentrate on enhancing the retriever's efficiency \cite{karpukhin2020dense, izacard2021contriever, chuang2023expand}, whereas others prioritize the reader's performance \cite{izacard2021leveraging, yu2022kgfid}. Additionally, there are studies that delve into the incorporation of structured knowledge from the knowledge base into the question answering system, such as \cite{oguz2022unikqa, yu2023decaf}, while the latter also utilizes logical form to improve the accuracy of answer generation. Although previous works are limited to small models with weaker reasoning capabilities, they have formed an efficient system of information retrieval, condensing the extracted knowledge corpus into brief statements. Therefore, our research integrates this system with LLMs to offer necessary knowledge and aid the LLMs in producing more dependable chains of thought.

Before LLMs, several works proposed methods for decomposing multi-hop questions into single-hop sub-questions \cite{min2019multihop, sun2020sparqa, khot2021text}. Some rule-based methods generate unnatural sub-questions, while other methods are constrained by the model's capacity. In contrast, our approach leverages CoT to induce LLMs to decompose complex multi-hop questions into sub-questions that are more comprehensible and logically coherent.

\section{Conclusion}
In this paper, we investigate the faithful reasoning of large language models on knowledge-intensive KBQA tasks. We propose a Knowledge-Driven Chain-of-Thought framework to improve the reasoning performance of large language models. Through experiments on knowledge-intensive KBQA tasks, we show that KD-CoT leads to superior performance with interpretable inference steps. We also present a CoT collection on the KBQA datasets that can be utilized for CoT fine-tuning and few-shots in-context learning. Additionally, we investigate a new training approach to develop a robust retriever that can efficiently access external knowledge, which results in a substantial improvement in the Hit scores for retrieved knowledge.

\section{Discussion \& Future work}
% 我们同时prompt模型生成检验子步骤的后续思维链以便比较修正前后的推理表现，后续工作可以限制模型仅生成到当前步骤的思维链，在经过与外部系统的交互后再进一步生成后续的推理步骤
Despite the fact that our method can efficiently access external knowledge and correct the sub-answers generated by the large language model, the final performance of LLM still leaves behind the current SOTA. This could be caused by: 1) The inability of the QA system to generate precise answers for all sub-questions, as our simply designed reader achieves only 73.7 Hit@1 on the WebQSP dataset. 2) The reasoning hallucination. LLM can still hallucinate reasoning despite our corrections to precise answers. Therefore, future work can focus on training a more robust reader or supervising both the reasoning questions and answers. Additionally, although we use greedy decoding for the QA system to generate candidate answers, our method is still costly as the entire framework contains an extra LLM and a verifier. In future work, more efficient techniques such as searching on the web or filtering retrieved knowledge can be utilized to further decrease the cost of KD-CoT.

% \section{Discussion \& Future work}
% % 我们同时prompt模型生成检验子步骤的后续思维链以便比较修正前后的推理表现，后续工作可以限制模型仅生成到当前步骤的思维链，在经过与外部系统的交互后再进一步生成后续的推理步骤
% Despite the fact that our method can efficiently access external knowledge and correct the sub-answers generated by the large language model, the final performance of LLM still leaves behind the current SOTA. This could be caused by: 1) The inability of the QA system to generate precise answers for all sub-questions, as our simply designed reader achieves only 73.7 Hit@1 on the WebQSP dataset. 2) The reasoning hallucination. LLM can still hallucinate reasoning despite our corrections to precise answers. Therefore, future work can focus on training a more robust reader or supervising both the reasoning questions and answers. Additionally, although we use greedy decoding for the QA system to generate candidate answers, our method is still costly as the entire framework contains an extra LLM and a verifier. In future work, more efficient techniques such as searching on the web or filtering retrieved knowledge can be utilized to further decrease the cost of KD-CoT.

% \clearpage

\bibliography{aaai24}

\clearpage

\section{Appendix}

\subsection{Implementation details}
We present here in detail the parameter settings we used to train the QA system and to perform CoT fine-tuning.
\begin{table}[ht]
    \centering
    \begin{minipage}[b]{1.0\linewidth}
    \centering
        \scalebox{0.95}{
        \begin{tabular}{l|cc}
            \toprule
            Model &  \# Params & \# Total Params \\
            BERT-base-uncased  & 110M & 110M \\
            T5-large & 770M & 770M \\
            Llama2-7b\_lora & 12M & 7B \\
            \hline
            T5-3B & 3B & 3B \\
            flan-T5-3B & 3B & 3B \\
            Llama2-7b\_lora & 12M & 7B \\
            \bottomrule
        \end{tabular}}
        \caption{Models utilized and their parameters. \# Params represents trainable parameters.}
    \end{minipage}
    \begin{minipage}[b]{1.0\linewidth}
    \centering
        \scalebox{0.8}{
        \begin{tabular}{l|ccc|cc}
            \toprule
            {} &  BERT & T5-large & Llama-7B & T5-3B & Flan-T5-3B\\
            \hline
            Lr & 2e-5 & 5e-5 & 1e-4 & 1e-4 & 1e-4 \\
            Batch Size & 128 & 16 & 32 & 32 & 32 \\
            Epoch & 40 & - & 5 & 5 & 5 \\
            Clip Norm & 2.0 & 1.0 & 1.0 & 1.0 & 1.0 \\
            \bottomrule
        \end{tabular}}
         \caption{Hyper-params settings for training QA system and for CoT fine-tuning.}
    \end{minipage}
    \label{hyper}
\end{table}

Specifically, we train our model using Deepspeed, with pytorch==2.0.0, peft==0.2.0, and transformers==4.29.1. For models larger than 1B, we train with precision bfloat16. We adopt LoRA and insert low-rank adapters with dimensions equal to 16 for fine-tuning Llama. For the optimizer and learning scheduler, we apply AdamW with Beta=[0.9, 0.95] and LinearDecay with a warmup ratio equal to 0.1. All experiments are conducted on 8 $\times$ 40G Nvidia A100. 
    
% \begin{figure}[ht]
%     \vspace{-3mm}
%     \centering
%     \includegraphics[width=1.0\linewidth]{input_inference.pdf}
%     \caption{(b) Input for LLM inference.}
%     \label{input_inference}
%     \vspace{-3mm}
% \end{figure}

\subsection{Demonstration illustration}
\begin{figure}[h]
    % \vspace{-2mm}
    \centering
    \includegraphics[width=1.0\linewidth]{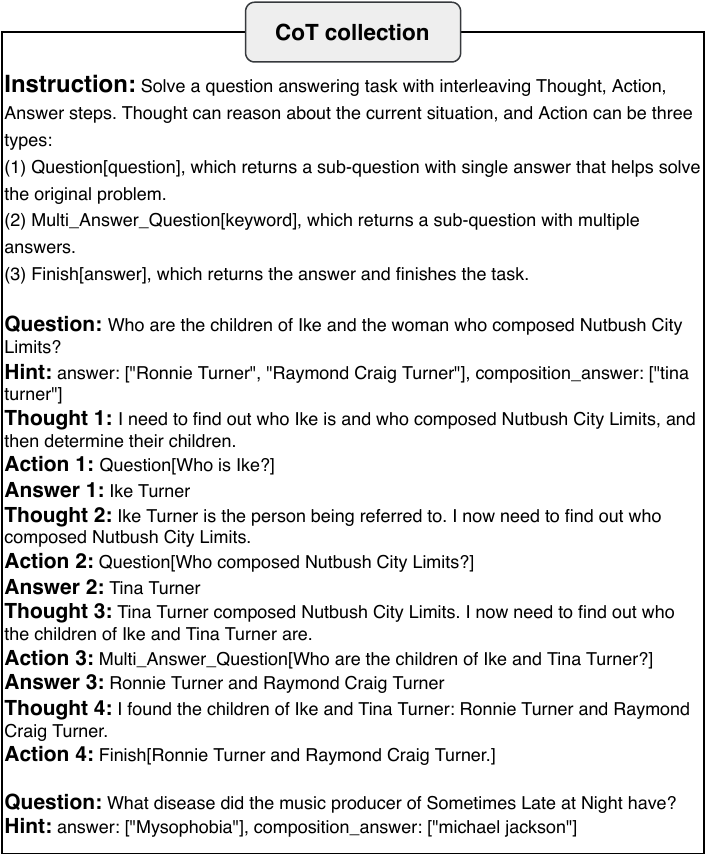}\\
    \includegraphics[width=1.0\linewidth]{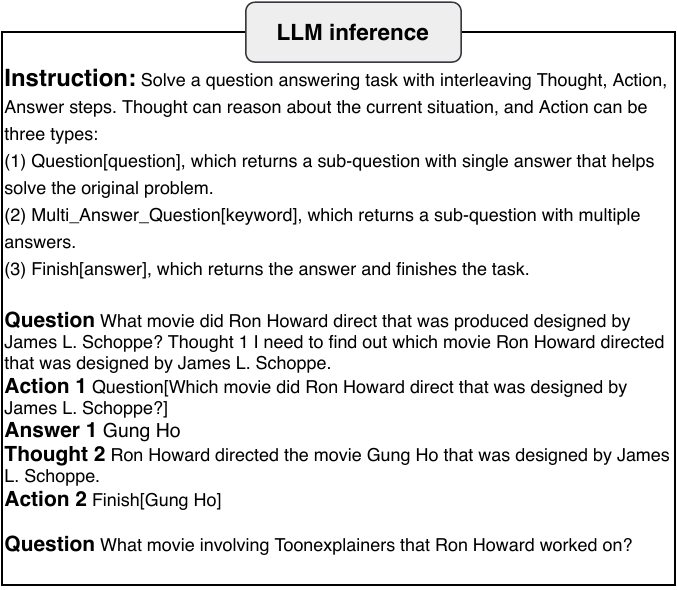}
    \caption{Input for constructing CoT collection (up) and input for LLM inference (down).}
    \label{input_cot}
\end{figure}

We add final answers and composition answers as "Hint" before structured rationale to construct our CoT collection. When conducting in-context learning during inference, we eliminate the "Hint" from the chosen demonstration. We provide two examples as illustrated in Figure \ref{input_cot}, one for constructing the CoT collection and the other for performing ICL on large language models during inference.

% \clearpage
\subsection{Case Analysis}
\begin{figure}[h]
    \centering
    \includegraphics[width=1.0\linewidth]{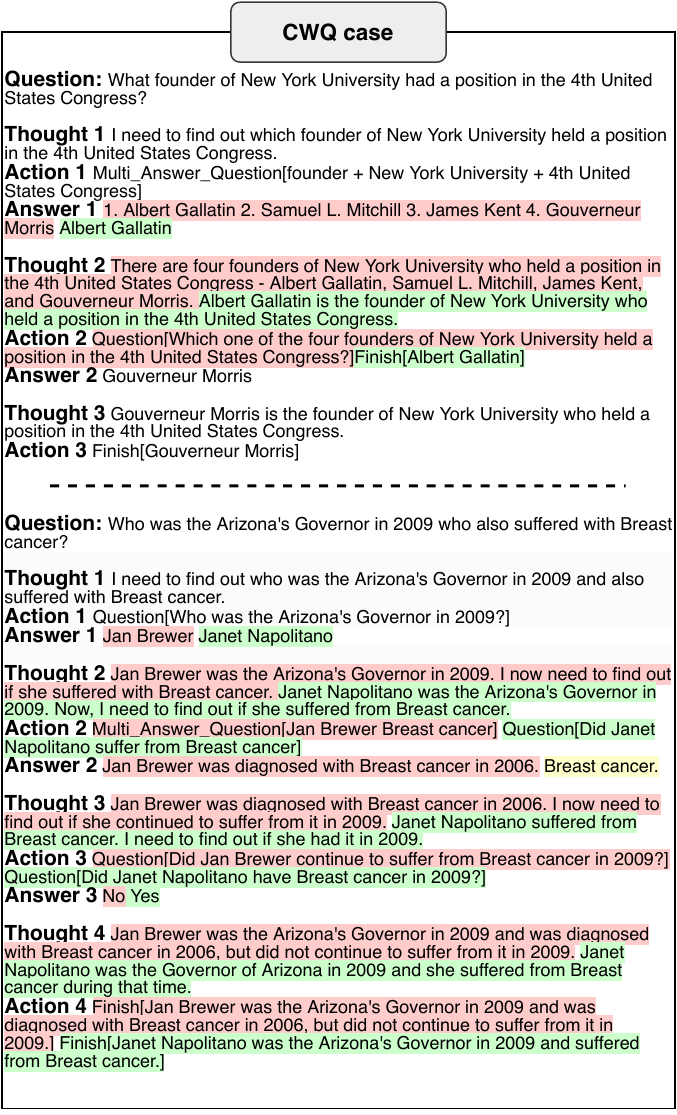}
    \caption{Cases of CWQ. Red and Green blocks represent the original hallucinations of LLM and the faithful reasoning after sub-answer correction. Yellow block signifies that the answer generated by the QA system is not entirely precise, but it does not impact the inference of the subsequent models.}
    \label{cwq_case}
\end{figure}

\begin{figure}[h]
    \centering
    \includegraphics[width=1.0\linewidth]{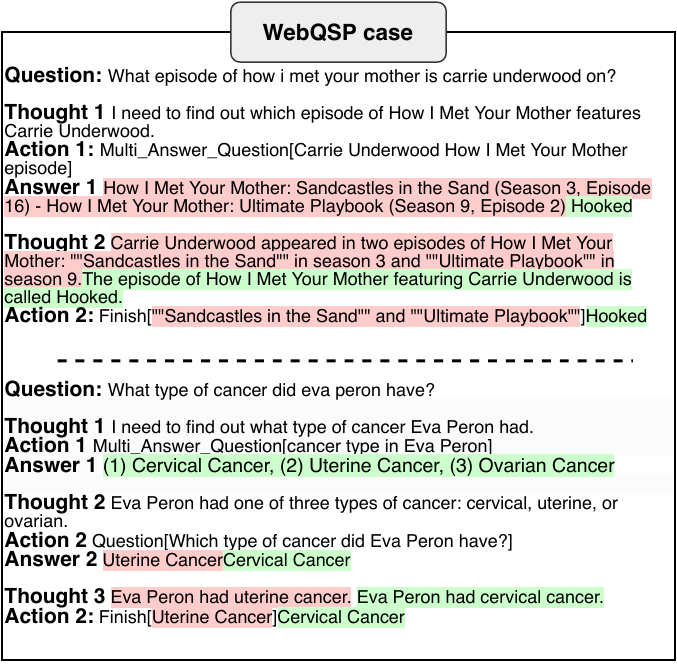}
    \caption{Cases of WebQSP, by applying KD-CoT we rectify the sub-answers and make the reasoning of LLM more faithful. Red and Green blocks represent the original hallucinations of LLM and the faithful reasoning after sub-answer correction.}
    \label{webqsp_case}
\end{figure}

We present in Figure \ref{cwq_case} and \ref{webqsp_case} several cases before and after our Knowledge-Driven Chain-of-Thought. We observe indicates that by correcting the sub-answers, the subsequent reasoning steps become more dependable and accurate, ultimately resulting in the correct final answers.

\end{document}